\documentclass[10pt,technote,cspaper,compsoc]{IEEEtran}
\ifCLASSOPTIONcompsoc
  \usepackage[nocompress]{cite}
\else
  \usepackage{cite}
\fi

\ifCLASSINFOpdf
\else
\fi

\usepackage{times}
\usepackage{epsfig}
\usepackage{graphicx}
\usepackage{amsmath}
\usepackage{amssymb}

\usepackage{footnote}
\usepackage{multirow}
\usepackage{booktabs}
\usepackage{threeparttable}
\usepackage{enumerate}
\usepackage{indentfirst}
\usepackage{array}
\usepackage{rotating}
\usepackage{ragged2e}
\usepackage{wasysym}
\usepackage{caption}

\def\ie{{\em i.e.}}
\def\eg{{\em e.g.}}
\def\etal{{\em et al.}}

\begin{document}

\title{Learning from Large-scale Noisy Web Data with Ubiquitous Reweighting for Image Classification}

\author{Jia~Li,~\IEEEmembership{Senior Member,~IEEE}, Yafei Song, Jianfeng Zhu, Lele Cheng, Ying Su, Lin Ye, Pengcheng Yuan and~Shumin Han
\IEEEcompsocitemizethanks{\IEEEcompsocthanksitem J. Li and P. Yuan are with the State Key Laboratory of Virtual Reality Technology and Systems, School of Computer Science and Engineering, Beihang University, Beijing, 100191, China. E-mail: \{jiali, yuanpengcheng\} @buaa.edu.cn. \protect
\IEEEcompsocthanksitem J. Li is also with the Shenzhen Cyberspace Laboratory, Shenzhen, China. \protect
\IEEEcompsocthanksitem Y. Song is with the National Engineering Laboratory for Video Technology, School of Electronics Engineering and Computer Science, Peking University, Beijing 100871, China. E-mail: songyf@pku.edu.cn.\protect
\IEEEcompsocthanksitem J. Zhu, L. Cheng, Y. Su, L. Ye and S. Han are with the Computer Vision Technology Department of Baidu, Beijing 100871, China. S. Han is the corresponding author. E-mail: \{zhujianfeng03, chenglele, suying02, yelin02, hanshumin\} @baidu.com \protect
}

}

\markboth{ }%
{Li \MakeLowercase{\textit{et al.}}: Learning from Large-scale Noisy Web Data with Ubiquitous Reweighting for Image Classification}

\IEEEtitleabstractindextext{%
\begin{abstract}
\justifying
Many advances of deep learning techniques originate from the efforts of addressing the image classification task on large-scale datasets. However, the construction of such clean datasets is costly and time-consuming since the Internet is overwhelmed by noisy images with inadequate and inaccurate tags. In this paper, we propose a Ubiquitous Reweighting Network (URNet) that learns an image classification model from large-scale noisy data. By observing the web data, we find that there are five key challenges, \ie, imbalanced class sizes, high intra-classes diversity and inter-class similarity, imprecise instances, insufficient representative instances, and ambiguous class labels. To alleviate these challenges, we assume that every training instance has the potential to contribute positively by alleviating the data bias and noise via reweighting the influence of each instance according to different class sizes, large instance clusters, its confidence, small instance bags and the labels. In this manner, the influence of bias and noise in the web data can be gradually alleviated, leading to the steadily improving performance of URNet. Experimental results in the WebVision 2018 challenge with 16 million noisy training images from 5000 classes show that our approach outperforms state-of-the-art models and ranks the first place in the image classification task.
\end{abstract}


}

\maketitle

\IEEEdisplaynontitleabstractindextext

%
\IEEEpeerreviewmaketitle

\section{Introduction}
\label{sec:intro}

\IEEEPARstart{I}{n} the past decade, the deep learning technique has achieved great advances under the assistance of large-scale visual datasets. Among these advances, many of highly influential innovations originate from the efforts in addressing the image classification task on ImageNet~\cite{ImageNetIJCV2015}, such as the multi-layer convolution structure of AlexNet~\cite{AlexNetNIPS2012}, very deep architecture of VGGNet~\cite{VGGNetICLR2015}, multi-scale inception of GoogleNet~\cite{GoogLeNetCVPR2015}, residual connections of ResNet~\cite{ResNetCVPR2016}, group convolution in ResNeXt~\cite{ResNeXtCVPR2017} and squeeze-and-excitation module in SENet~\cite{SENetCVPR2018}. Based on these innovations in image classification, the deep learning technique has also obtained remarkable success in many other applications such as autonomous driving~\cite{DeepDrivingICCV2015}, robotic control~\cite{DLRoboticsIJRR2018}, medical imaging diagnostics~\cite{DLMedicalMIA2017} and game playing~\cite{GamePlayingNature2015}.

With the usage of these deep learning techniques, state-of-the-art models have achieved impressive high accuracy in the image classification task on ImageNet. The model proposed by He \etal~\cite{He_2015_ICCV} even outperforms human. The improvement subsequently becomes marginal in recent years, which has leaded to the termination of this challenge from 2017. However, quite a few researchers still think that the image classification problem has not been completely solved, since ImageNet is a clean and well annotated dataset, while the data in real-world are usually bias and noisy. Human can learn exact knowledge from such data, while state-of-the-art deep models may not. Therefore, researchers are now looking urgently at the datasets which are more similar to real-world data. To this end, Li \etal~\cite{WebVisionArXiv2017} have constructed and published the WebVision dataset, which currently consists of $16$ million images within five thousand classes. For each class, the images are retrieved from Google and Flickr using specific retrieval words. The retrieval words are also taken as the label of this class. By means of modern search engine, such large-scale image dataset can be constructed quickly as well as with low cost. However, the retrieved results are not always accurate, the algorithms as well as the corresponding models will face some key challenges to exploit such a dataset. To accelerate the research progress on these challenges, Li~\etal~\cite{WebVisionArXiv2017} have organized the WebVision Image Classification Challenge since 2017.

To effectively exploit the large-scale noisy dataset, current works usually follow two types of methodologies. Some works focus on purifying the noisy dataset. For example, Veit~\etal~\cite{veit2017learning} resorted to a clean sub-set and the multi-task learning algorithm to jointly clean noisy annotations and accurately classify images. Wang~\etal~\cite{wang2018iterative} proposed an iterative learning framework to detect the noisy samples with open-set labels so as to pull away noisy samples from clean samples. Tanaka~\etal~\cite{tanaka2018joint} proposed a joint optimization framework to update the parameters and estimate true labels. These methods aim to learn the accurate model as well as clean the dataset. However, it is a more difficult task to correct the label of each sample, which also may be not necessary.

The  other works focus on decreasing the weight of noisy samples so as to alleviate their negative effects. For example, Fu~\etal~\cite{fu2015relaxing} assumed that the features of true positive samples are similar and the features of noisy samples are variant, which is also an important achievement for deep learning methods to tackle the selectivity-invariance dilemma \cite{lecun2015deep}. Based on this assumption, they minimized the discrepancy between the affinity representation and its low rank approximation to decrease the influence of noisy samples. Zhuang~\etal~\cite{zhuang2017attend} randomly grouped multiple images into a single training instance and adopted attention strategy to suppress the noisy signal. Niu~\etal~\cite{Niu_2018_CVPR} adopted more accessible category-level supervision. In particular, they built their deep probabilistic framework upon variational autoencoder (VAE), in which the classification network and VAE can jointly leverage category-level hybrid information. These methods focus on reweighting the noisy samples, however, both noise and bias are ubiquitous.

With previous works in mind, we carefully analyze the WebVision dataset, and find that there may be five primary difficulties during exploiting the large-scale noisy web data. First, the size of each class varies from about three hundred to fifty thousand, which leads to imbalanced class sizes. Second, as some classes are really similar to each other, these classes usually have high intra-classes diversity and inter-class similarity. Third, since the instances are directly retrieved from Internet, quite a few instances are with imprecise or wrong labels. Fourth, the common instances may be overwhelmed by uncommon instances, therefore representative instances are insufficient. Last but not least, due to the ambiguous class labels, some instances may actually belong to several classes at the same time.

To address these challenges, we assume that every instance has the potential to contribute positively by alleviating the corresponding data noise via reweighting the influence of different classes as well as their labels, large instance clusters, small instance bags and every instance. In this manner, the influence of noise in the web data can be gradually alleviated, leading to the steadily improving performances of the proposed URNet model. Experimental results in the WebVision 2018 challenge with 16 million noisy training images from 5000 classes show that our approach outperforms state-of-the-art models and ranks the first place in the image classification task. The contributions of this paper mainly lie in three aspects:
\begin{enumerate}
\item We thoroughly analyze the challenges while exploiting the large-scale noisy web data and summarize five potential causes, which may inspire further studies. 
\item We propose a ubiquitous reweighting network as well as five reweighting strategies which are adopted to alleviate the imbalanced class sizes, high intra-classes diversity and inter-class similarity, imprecise instance labels, insufficient representative instances, and ambiguous class labels. 
\item With the proposed URNet, we can exploit the large-scale noisy web data effectively to perform image classification, and achieve the first place in the WebVision 2018 challenge. 
\end{enumerate}


The rest of this paper is organized as follows. In Section~\ref{sec:related:work}, we briefly review related works. In Section~\ref{sec:challenge}, we summarize five key challenges to exploit large-scale noisy data. Accordingly, our URNet along with five reweighting strategies are detailedly described in Section~\ref{sec:approach}. Finally, we show the experimental results in Section~\ref{sec:experiments} and conclude this paper in Section~\ref{sec:conclusion}.

\section{Related Work}
\label{sec:related:work}

Benefitting from the carefully annotated big visual dataset and powerful convolutional neural networks, researchers have achieved great progresses recently on the image classification problem. In 2012, Alex \etal~\cite{AlexNetNIPS2012} first used GPU to accelerate the time-consuming training process of a deep CNN model and obtained a great improvement than previous shallow models, This work demonstrates the powerful representative ability of a deep model when proessing images. After that, it is generally assumed that deeper and wider models would obtain better performance due to their more powerful representative ability. Some works then focus on increasing the depth as well as the width. For example, Simonyan~\etal~\cite{VGGNetICLR2015} proposed the VGGNet and successfully trained it, which extends the depth of a single model from several weight layers to $19$. This work breaks previous depth limit and demonstrates that it is possible to effectively construct and train a very deep model. After that, He~\etal~\cite{ResNetCVPR2016} proposed the residual structure as a block module to construct the ResNet. Benefitting from this structure, gradient can directly flow from high layers to low layers. This technology makes it possible to effectively train an ultra-deep model with $152$ weight layers. Moreover, the model doesn't suffer seriously difficulties even with over one thousand weight layers. He~\etal~\cite{ResNeXtCVPR2017} further proposed ResNeXt to improve the residual block. Besides the depth, some other works focus on the width. For example, Szegedy~\etal~\cite{GoogLeNetCVPR2015} proposed multi-scale inception module and construct the GoogleNet, which can encode multi-scale features in one layer. Huang~\etal~\cite{DenseNet_Huang_2017_CVPR} proposed the densely connected networks, in which each layer connects with all its previous layers. Based on these works, some improved networks~\cite{hu2017squeeze, xie2017aggregated, chen2017dual} were also proposed to achieve higher accuracy or computational efficiency. These methods usually focus on the image classification problem on carefully annotated big visual dataset and pay less attention on the noisy or bias dataset.

To effectively exploit the large-scale noisy dataset, current works usually follow two types of methodologies. Some works focus on purifying the noisy dataset. These methods may resort to a clean sub-set~\cite{veit2017learning}, detect the noisy samples~\cite{wang2018iterative} and try to correct the wrong labels~\cite{tanaka2018joint}. Specifically, Veit~\etal~\cite{veit2017learning} resorted to a clean sub-set and multi-task learning to jointly learn to clean noisy annotations and to accurately classify images. This work demonstrates how to use the clean annotations to reduce the noise in the large-scale dataset before fine-tuning the network using both the clean set and the full set with reduced noise. Wang~\etal~\cite{wang2018iterative} proposed an iterative learning framework to detect the noisy samples with open-set labels so as to pull away noisy samples from clean samples. To benefit from the noisy label detection, they designed a Siamese network to encourage clean labels and noisy labels to be dissimilar. A reweighting module is also applied to simultaneously emphasize the learning from clean labels and reduce the effect caused by noisy labels. Tanaka~\etal~\cite{tanaka2018joint} proposed a joint optimization framework of updating parameters and estimating true labels. They optimized the labels themselves instead of treating the noisy labels as fixed. The joint optimization of network parameters and the noisy labels can correct inaccurate labels as well as improve the performance of the classifier. This group of methods aims to learn the accurate model as well as correct the noisy labels. However, it is a more difficult task to correct the noisy labels simultaneously than to only learn an accurate model, which also may be not necessary.

The other works focus on decreasing the weight of noisy samples so as to alleviate their negative effects. These methods usually resort to the similarity between intra-classes features~\cite{fu2015relaxing}, multi-instance learning, attention network~\cite{zhuang2017attend}, or category-level supervision~\cite{Niu_2018_CVPR}. Specifically, Fu~\etal~\cite{fu2015relaxing} assumed that the features of true positive samples in a class are similar and the features of noisy samples are very different. They embedded the feature map of the last deep layer into a new affinity representation, and further minimized the discrepancy between the affinity representation and its low-rank approximation.
Zhuang~\etal~\cite{zhuang2017attend} proposed two unified strategies, \ie, random grouping and attention, to reduce the negative impact of noisy annotations. Specifically, random grouping is to stack multiple images into a single training instance and increase the labeling accuracy at the instance level. Attention, on the other hand, is to suppress the noisy signals from both incorrectly labeled images and less discriminative image regions. This method can increase the labeling accuracy at the instance level and effectively reduce the negative impact of noisy annotations.  Niu~\etal~\cite{Niu_2018_CVPR} adopted a more accessible category-level supervision strategy. In particular, they built a deep probabilistic framework upon variational autoencoder (VAE), in which classification network and VAE can jointly leverage category-level hybrid information. Then, the method was extended for domain adaptation followed by a low-rank refinement strategy.
This group of methods focuses on reweighting the noisy samples, however, besides noise, data bias are also ubiquitous in large-scale real-world data.

\section{Challenge Analysis}
\label{sec:challenge}

To motivate the algorithm design, we analyze the large-scale noisy web data, \ie~the WebVision dataset, and summarize five primary difficulties. These difficulties are not specific only for WebVison but generally exist in the real-word data. Based on these analysis, we propose the ubiquitous reweighting strategy as well as the URNet. We also expect that our analysis could inspire the researchers to design more specific algorithms so as to exploit the large-scale noisy real-world data effectively.

Firstly, as is well-known, the Pareto principle extensively exists and the distribution of real-world things has high imbalance. To be specific, the Internet users do not fairly share each class of images, and the number of images in each class varies in a really wide range. The number of images is large in a common class, while it is small in a uncommon class. Actually, the number of images in a class varies from about three hundred to more than ten thousand. If we assign the same weight for each instance, the model will pay more attention on the class with more instances. And the class with small size will be overwhelmed by the class with large size. To alleviate this problem, the straight-forward method is to random sample a subset from the large class. However, this method can only make use of partial data. In this paper, we make each class to have almost the same influence on the model via reweighting each instance. Then the model can pay equal attention on each class while exploit all of the instances.

\begin{figure}
  \includegraphics[width=\linewidth]{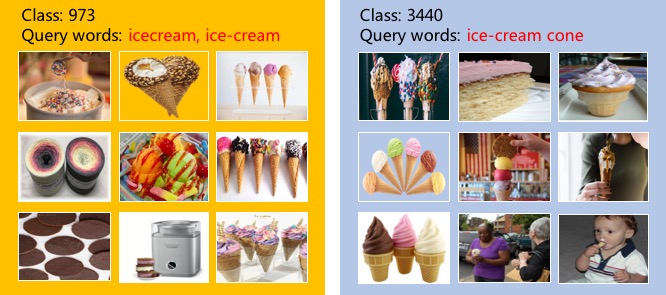}
  \caption{The data have high intra-classes diversity and inter-class similarity, \eg~in the class 973 and 3440, instances in one class are very different from each other, while some instances belonging to different classes are quite similar.}
  \label{fig:cluster:problem}
\end{figure}

Secondly, the instances usually have high intra-classes diversity and inter-class similarity. As shown in Fig.~\ref{fig:cluster:problem}, due to the various forms, shapes, illuminations, appearance and views, the instances in the same class may be extremely different from each other. However, these instances have the same semantic concept. On the other hand, the instances in one class may be really similar to the instances in another class due to their similar semantic concept, \eg~the class 973 and 3440 in Fig.~\ref{fig:cluster:problem}. Due to this phenomenon, some instances are really difficult to tackle. Intuitively, these difficult instances should be paid more attention by the model. To this end, we propose to cluster all the instances according to their appearance similarity. Intuitively, the instances in the large cluster may be similar to more instances from other classes, which should be paid more attention. That is to say, each instance can be reweighted by its cluster size. In this manner, we can automatically detect the difficult instances and reweight them.

Thirdly, the instances in the dataset are directly retrieved from the image websites, \ie~google and flickr, using query words. The query words are taken as the semantic label. Although these modern image search engines have made great progresses recently, there are still considerable amount of inaccurate instances, which are noise for the dataset and would confuse the model. To suppress their negative influence, we can detect them or may further correct them. However, this process is more difficult and may also introduce additional noise. In this paper, we adopt a more flexible method and estimate the confidence of each instance, which is then used to reweight the instance. More confidence the instance is, more weight it has.

\begin{figure}
  \includegraphics[width=\linewidth]{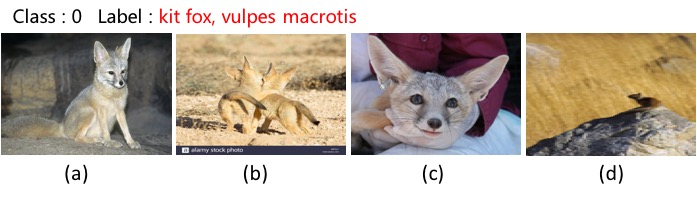}
  \caption{Insufficient representative instances, \eg~image (a) is an representative instance and has more information than others, but may be overwhelmed by the uncommon instances like (b), (c), (d).}
  \label{fig:high:diversity}
\end{figure}

Fourthly, in one specific class, some instances have more information while some others have less, \eg~in Fig.~\ref{fig:high:diversity}, the first image may contain more information than the others. We refer the instances like Fig.~\ref{fig:high:diversity}(a) as representative instances. However, the instances in one class usually have a very high diversity of appearance, pose, lightness, and so on. This leads to insufficient representative instances, which may be overwhelmed by uncommon instances. To alleviate this problem, we resort to the bag-specific instance saliency technology to discover the representative instances and highlight them.

\begin{figure}
  \center
  \includegraphics[width=0.8\linewidth]{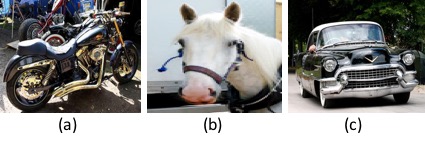}
  \caption{One instance may belong to several classes simultaneously due to the ambiguous class labels, \eg~image (a) may belong to class 2334: motorcycle policeman, speed cop, motorcycle cop (id: label), and class 3369: trail bike, dirt bike, motorcycle bike, image (b) may belong to class 0331: seed, conker seed, and class 1044: Equus caballus, female horse, horse, image (c) may belong to class 0268: jeep, landrover, and class 0563: paddle wheel, paddlewheel.}
  \label{fig:ambiguity:label}
\end{figure}

Last but not least, due to the semantic confusion and ambiguous class labels, all the classes are not orthogonal to each other. This means that some instances may actually belong to several classes at the same time, \eg~the images in Fig.~\ref{fig:ambiguity:label}. The model will be confused to tackle these instances. To alleviate this problem, we resort to a smoothing label as the optimization target rather than a hard label. With the smoothing label, one instances can be regarded as belonging to several classes at the same time.

\section{Ubiquitous Reweighting}
\label{sec:approach}

Based on our analysis, we propose a ubiquitous reweighting network to alleviate the five challenges. Specifically, to handle imbalanced classes, a class-weighted loss is defined, which is to restrain the effect of classes with large size, while enhance the effect of the classes with small size. To tackle the high intra-classes diversity and inter-class similarity, we designed a cluster-weighted strategy. With this strategy, the instances in confusable clusters are further emphasized. To alleviate the imprecise instances, a model is trained to predict the correlation score of each instance relative to its query words. Each instance is reweighted according to its correlation score. The higher correlation score, the higher weight. In consideration of insufficient representative instances, the instances are organized into small bags. The salient instance in a bag usually contain more information, which is reweighted via an attention network. To alleviate the ambiguous class labels, not only the ground truth label but also the label predicted by the baseline model are both used to formulate a joint loss function.

\subsection{Reweighting Class}
\label{sec:reweighting:classes}

To perform the image classification task, we construct our URNet based on ResNeXt-101~\cite{ResNeXtCVPR2017}, which is among state-of-the-art image classification models. The detailed structure of the network can be found in \cite{ResNeXtCVPR2017}. The only difference lies in the output layer, which is modified to $C = 5, 000$ dimensions, which are corresponding to the $5, 000$ classes in the WebVision dataset. We first train the model via minimizing the standard cross entropy loss
\begin{equation}
\mathcal{L}_{sce} = - \sum_{i=1}^{N} \sum_{j=1}^{C} {y_{ij} \log(p_{ij})},
\label{eq:cross:entropy}
\end{equation}
where $N$ is the total number of instances in the dataset, $y_{ij} \in \{0, 1\}$ and $y_{ij} = 1$ only if the instance $\mathcal{I}_{i}$ is belong to the $j$-th class $\mathcal{C}_j$, $p_{ij}$ is the probability of $\mathcal{I}_{i} \in \mathcal{C}_j$ which is predicted by the model. This trained model is taken as the baseline model.

From the standard cross entropy loss~\eqref{eq:cross:entropy}, we can see that each instance contributes equally to the loss function. However, if the imbalanced class size is taken into consideration, the classes with a few instances will be overwhelmed by the classes with a lot of instances. To alleviate this problem, the standard cross entropy loss is then modified to the class weighted loss as
\begin{equation}
\mathcal{L}_{wce} = - \sum_{i=1}^{N} \sum_{j=1}^{C} w_j^{\mathcal{C}} {y_{ij} \log(p_{ij})},
\label{eq:cross:entropy:class:weight}
\end{equation}
where $w_j^{\mathcal{C}}$ is the weight of each class, which can be simply defined as inversely proportional to the class size
\begin{equation}
w_{j}^{\mathcal{C}} = \dfrac{r_j}{\sum_{j = 1}^{C} r_j}, \text{where}~r_j = \dfrac{1}{N_j}.
\label{eq:class:weight}
\end{equation}
In this way, each class has the same weight to the loss function. Furthermore, to be more flexible, a parameter $\alpha \in [0, 1]$ is introduced to adjust from equal instance weight to equal class weight, then the class weight
\begin{equation}
w_j^{\mathcal{C}} = (1 - \alpha) + \alpha \dfrac{r_j}{\sum_{j = 1}^{C} r_j}.
\label{eq:class:weight:soft}
\end{equation}
When $\alpha = 0$, each instance contributes equally to the loss, while when $\alpha = 1$, each class contributes equally to the loss. In our experiments, we set $\alpha = 0.5$ empirically.

\subsection{Reweighting Cluster} 
\label{sec:reweighting:cluster}

The large-scale noisy data have high intra-classes diversity and inter-class similarity, which makes it more difficult to correctly classify the instances, especially for them which are similar to the instances from other classes. To alleviate this issue, we design a reweighting strategy based on unsupervised clustering to increase the weight of such instances.


We first evaluate the inter-class similarity, and a confusion matrix $\mathcal{M}$ is constructed for all classes, where an element $\mathcal{M}_{jk}$ records the number of instances which are belong to the $j$-th class but classified as the $k$-th class by the baseline model. For a class $\mathcal{C}_j$, its four most confused classes $\mathcal{C}_{jk_{1}}, \mathcal{C}_{jk_2}, \mathcal{C}_{jk_3}, \mathcal{C}_{jk_4}$ are selected according to the confusion matrix. Combined with the class $\mathcal{C}_{j}$, all instances in these five classes are unsupervisedly clustered into five groups $\mathcal{G}_{j1}, \mathcal{G}_{j2}, \mathcal{G}_{j3}, \mathcal{G}_{j4}, \mathcal{G}_{j5}$ using k-means algorithm \cite{hartigan1979algorithm}. Specifically, we take the output of the layer before the classification layer of the baseline model as the image features.

Due to the high intra-classes diversity and inter-class similarity, the instances from different classes may be assigned into a group only because of their similar appearance, and vice versa. Moreover, the group sizes usually vary in a wide range. It is obvious that one instance in a big group has similar appearance to many other instances. Such an instance is usually difficult for classification and should be paid more attention by the model. In other words, each instance should be reweighted according to its group size. Then the cluster weight is defined as
\begin{equation}
w_j^{\mathcal{G}_{j_l}} = \dfrac{N_{\mathcal{G}_{j_l}}}{\sum_{t = 1}^{5} N_{\mathcal{G}_{j_t}}}, l \in \{1, 2, 3, 4, 5\}
\end{equation}
where $N_{\mathcal{G}_{j_l}}$ is the number of instances in group $\mathcal{G}_{j_l}$. The cluster weight of each instance can be further defined as
\begin{equation}
w_{ij}^{\mathcal{G}} =
  \left\{
    \begin{array}{cl}
      w_j^{\mathcal{G}_{j_l}}, & \text{if}~\mathcal{I}_i \in \mathcal{C}_j \cap \mathcal{G}_{j_l} \\
      0, & \text{otherwise}
    \end{array}
  \right..
\end{equation}
Taking cluster weight into consideration, the weighted cross entropy loss \eqref{eq:cross:entropy:class:weight} can be modified as
\begin{equation}
\mathcal{L}_{wce} = - \sum_{i=1}^{N} \sum_{j=1}^{C} w_{ij}^{\mathcal{G}} w_j^{\mathcal{C}} {y_{ij} \log(p_{ij})}.
\label{eq:cross:entropy:cluster}
\end{equation}

Benefiting from the cluster weight, the model can pay more attention to the instances which are easily confused with other classes, and the performance is significantly improved. Moreover, as the clustering algorithm is unsupervised, there is no more manual work to be done.

\subsection{Reweighting Instance}
\label{sec:reweighting:instance}

There are many imprecise instances in the large-scale noisy data. To alleviate this problem, we predict the confidence of each instance and then reweight it accordingly. The more confidential the instance is, the higher weight it should have. A straight-forward solution is to train a model which aims to predict the confidence. However, the influence of imprecise instances is ubiquitous. It is a problem to construct the training set. By observing the data, we find that the top ranking images of each class from Google are usually high confidential. Therefore, we select the top $30$ ranking images along with their labels as the positive training samples. At the same time, for each image, we random select a label except its ground-truth label to construct the negative training samples.

\begin{figure*}[t]
\center
  \includegraphics[width=14cm]{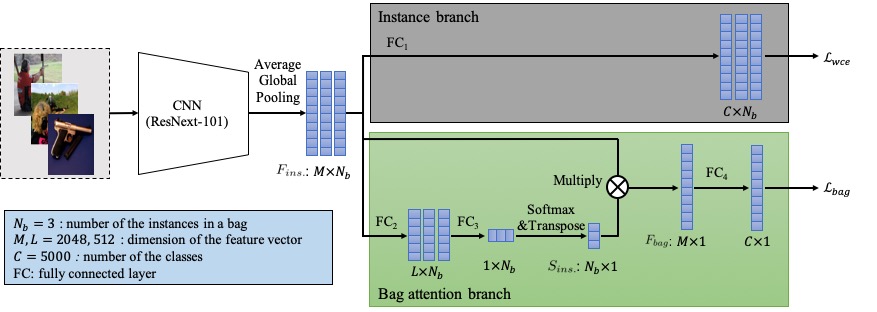}
  \caption{We add the bag attention branch to reweight bag-specific instance saliency. With this branch, the model can pay more attention on the representative instances.}
  \label{fig:bag:loss}
\end{figure*}

\begin{figure}[t]
\center
  \includegraphics[width=\linewidth]{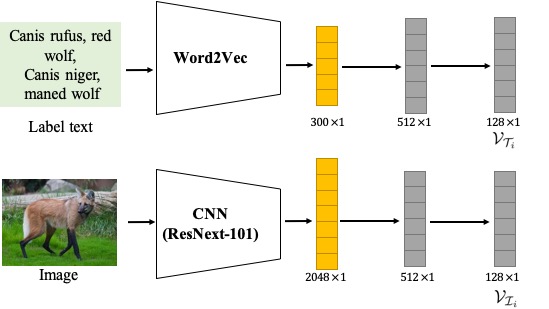}
  \caption{We take the correlation between the image features $\mathcal{V}_{\mathcal{I}_{i}}$ and the text features $\mathcal{V}_{\mathcal{T}_{i}}$ as the confidence score.}
  \label{fig:instance}
\end{figure}

Instead of constructing an regression model, we resort to construct the relation between image and text features~\cite{wang2016learning, Ryan2014uvse}. As shown in Fig~\ref{fig:instance}, our confidence model takes the correlation between the image features $\mathcal{V}_{\mathcal{I}_{i}}$ and the text features $\mathcal{V}_{\mathcal{T}_{i}}$ as the confidence score, where the text features are extracted from the words corresponding to the label, the image features are extracted using the baseline model. To extract features from the words, we adopt the fastText-300~\cite{fastTextTACLT2017} model, whose last layer is replaced by two fully connected layers and output text features $\mathcal{V}_{\mathcal{T}_{i}}$. The last layer of the baseline model is also replaced by two fully connected layer, which output image features $\mathcal{V}_{\mathcal{I}_{i}}$. Note that, the dimensions of feature $\mathcal{V}_{\mathcal{I}_{i}}$ and $\mathcal{V}_{\mathcal{T}_{i}}$ are the same. Then the model can be trained via minimizing the correlation loss
\begin{equation}
\mathcal{L}_{tic} = - \sum_{i=1}^{N'} ( \cos( \langle \mathcal{V}_{\mathcal{I}_{i}}, \mathcal{V}_{\mathcal{T}_{i}} \rangle) - \cos( \langle \mathcal{V}_{\mathcal{I}_{i}}, \mathcal{V}_{\widetilde{\mathcal{T}_{i}}} \rangle) ),
\label{eq:loss:correlation}
\end{equation}
where $N' = 30 \times C$,  $\widetilde{\mathcal{T}_{i}}$ is the text of a random other class which is different from $\mathcal{T}_{i}$. By minimizing this loss, we can maximize the correlation between the image features and its ground-truth text features, while minimize the correlation between the image features and other text features.
As soon as the model is trained, the correlation $w_{i}^{\mathcal{I}} = \cos( \langle \mathcal{V}_{\mathcal{I}_{i}}, \mathcal{V}_{\mathcal{T}_{i}} \rangle)$ can be used to evaluate the confidence of each instance in the dataset. Then the weighted cross entropy loss \eqref{eq:cross:entropy:cluster} can be modified as
\begin{equation}
\mathcal{L}_{wce} = - \sum_{i=1}^{N} \sum_{j=1}^{C} w_{i}^{\mathcal{I}} w_{ij}^{\mathcal{G}} w_j^{\mathcal{C}} {y_{ij} \log(p_{ij})}.
\label{eq:cross:entropy:instance}
\end{equation}

\subsection{Reweighting Bag-Specific Instance Saliency}
\label{sec:reweighting:positive:instance}

The noisy large-scale dataset have high intra-classes diversity, which leads to insufficient representative instances. To highlight such instances, we reweight each instance according to its saliency, as the saliency can evaluate the representativeness. The straight-forward method is to adopt attention network~\cite{yin2015abcnn, teh2016attention, wang2017residual}, which is also similar to multi-instance learning~\cite{wu2015deep, alpaydin2015single, li2015multiple}. To this end, we random take $N_b$ instances from one class as a bag, where $N_b = 3$. Each bag is assumed to have at least one representative instance. Even though only $30 \%$ instances are representative, this assumption still will be satisfied under more than $97 \%$ situations.

As shown in Fig.~\ref{fig:bag:loss}, besides the standard instance branch, we add a new branch named bag attention branch. Its input are the $N_b$ feature vectors extracted from the images in a bag respectively, which is denoted as $F_{ins.}$. In this branch, there are $2$ fully connected layers, \ie~$\text{FC}_\text{2}$ and $\text{FC}_\text{3}$, which can predict the saliency of all instances along with softmax normalization operation, which is denoted $S_{ins.}$. Then we can take the weighted summation $F_{ins.} \times S_{ins.}$ as the features of the bag, which is denoted as $F_{bag}$. Another fully connected layer $\text{FC}_\text{4}$ can further estimate the probability when the image is belonging to each class. We use the cross entropy loss function to train this branch as well as the backbone network. The bag attention branch can highlight the saliency instance and its features so as to enable the model to pay more attention on the representative instances.

\subsection{Reweighting Label}
\label{sec:reweighting:Label}

For the dataset, only one class is assigned for each instance. However, an instance may actually belong to several classes at the same time. These instances may confuse the model. To this end, we further reweight the loss \eqref{eq:cross:entropy:instance} with smoothing label as
\begin{equation}
\mathcal{L}_{wce} = - \sum_{i=1}^{N} \sum_{j=1}^{C} w_{i}^{\mathcal{I}} w_{ij}^{\mathcal{G}} w_j^{\mathcal{C}} { ( \beta y_{ij} + (1 - \beta) \hat{y}_{ij}) \log(p_{ij})},
\label{eq:loss:label}
\end{equation}
where $\hat{y}_{ij} \in \{0, 1\}$ and $ \hat{y}_{ij} = 1$ only if $\mathcal{I}_i$ is supposed to belong to class $\mathcal{C}_j$, the parameter $\beta$ is to balance the ground-truth label and the supposed label. In our experiments, we set $\beta = 0.8$ empirically.

To obtain the supposed label, we also resort to the baseline model. Actually, its predicted result is directly taken as the supposed label. Note that, when the predicted result is same with the ground-truth, the loss is unchanged. When the predicted result is different with the ground-truth, instead of trusting any one, we think that the instance may have some uncertainties and this loss is a trade-off.

\begin{figure*}[t]
\center
  \includegraphics[width=12cm]{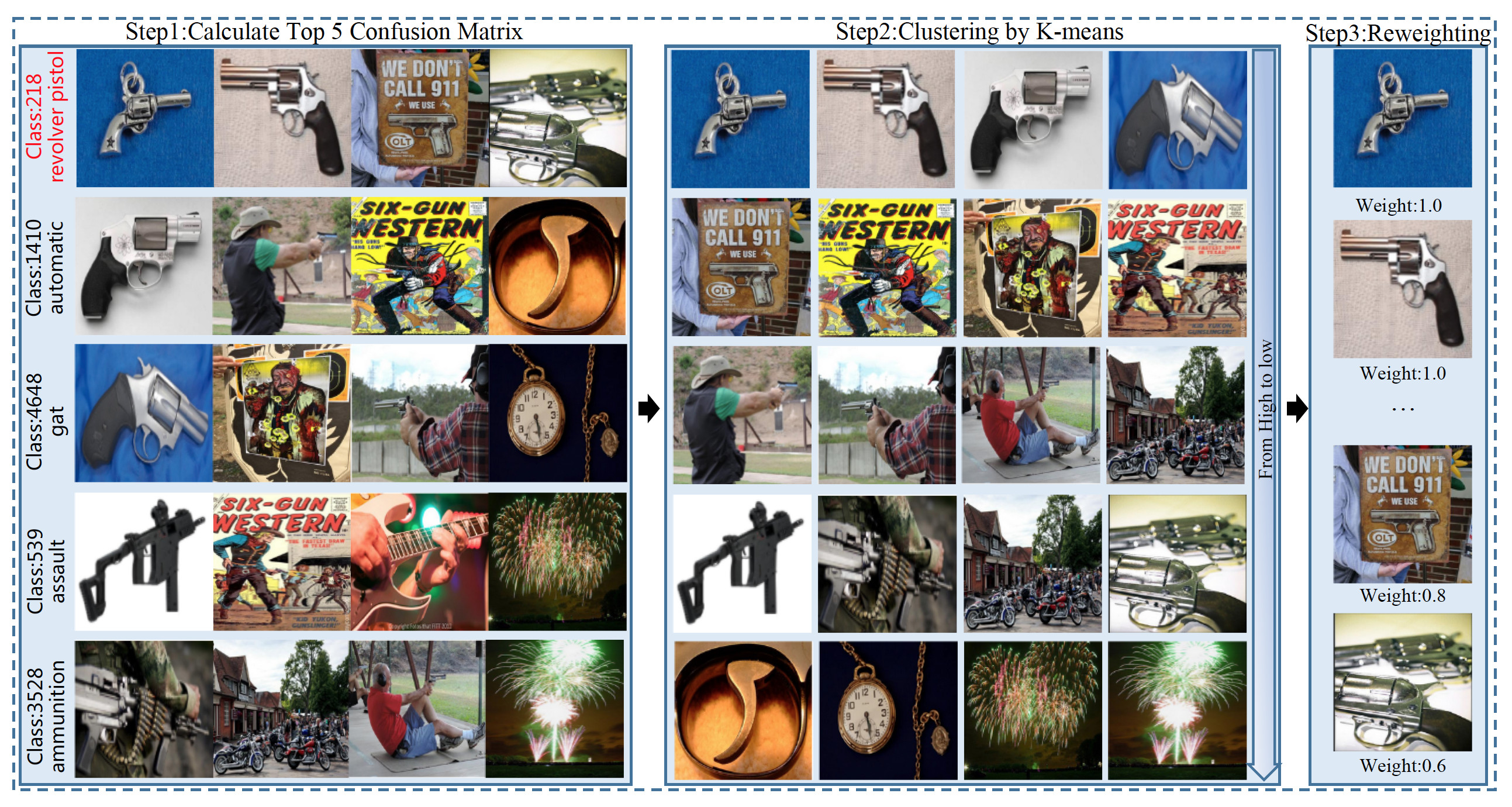}
  \caption{Some examples of clustering result on class 218 and its top-4 confusing classes.}
  \label{fig:cluster:result}
\end{figure*}

\begin{figure}[t]
  \includegraphics[width=\linewidth]{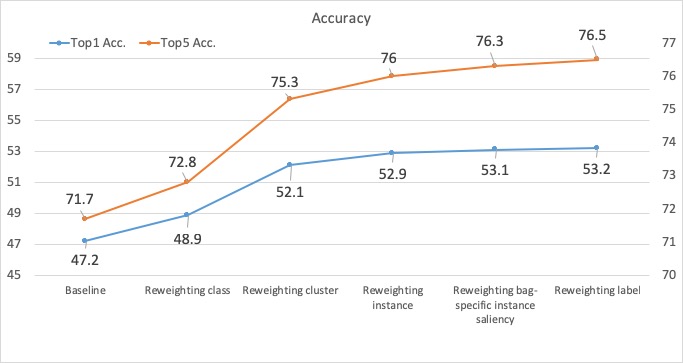}
  \caption{The effectiveness of each reweighting strategy. From the figure, we can see that the performance is improved steadily with our reweighting strategies.}
  \label{fig:acc:result}
\end{figure}

\section{Experiments}
\label{sec:experiments}

Benefiting from the ubiquitous reweighting network along with some other experimental tricks, our model ranks the first place at the WebVision image classification challenge 2018. In this section, we introduce the experimental details and verify the effectiveness of each module of our URNet.

In the experiments, we train our model using the training dateset published by WebVision, and evaluate it using the validation dataset. To train the baseline model ResNeXt-101, we take the previous model architecture designed for ImageNet and modify the output layer so as to make it corresponding to the $5, 000$ classes defined by WebVision dataset~\footnote{We only use the model architecture and the parameters are randomly initialized. The pretrained parameters of ResNeXt-101 on ImageNet are not used to follow the rules of WebVision 2018 Challenge. The expression of ``we take the pretrained model on ImageNet'' is a typo in the previous version~\cite{jialiwebvisionv1}.}. Then the model is fine-tuned by minimizing the cross entropy loss on the WebVision training set. The baseline model is then adopted to further construct the URNet step by step. And it is also taken to construct the assistant data needed by URNet, such as to construct the confusion matrix, predict the confidence of each instance.

\subsection{Effectiveness of Each Step}
\label{sec:details}

As shown in Fig.~\ref{fig:acc:result}, the baseline model achieves top-5 accuracy $71.7\%$. Based on the baseline model, we add each reweighting strategy one by one, \ie~reweighting class, reweighting cluster, reweighting instance, reweighting bag-specific instance saliency, reweighting label. Then the performance is improved steadily to $72.8\%, 75.3\%, 76\%, 76.3\%, 76.5\%$ respectively.

We can see that reweighting cluster is the most helpful strategy, which can increase the top-5 accuracy by $2.5\%$. As shown in Fig.~\ref{fig:cluster:result}, we demonstrate the clustering result of class 218 and its confusing four classes. We can see that, benefiting from this strategy, the model can pay more attention to the instances which are confusing with the instances from other classes. The large improvement also verifies the effectiveness of our reweighting cluster strategy.

The second important strategy is reweighting class. Class bias not only exists in the WebVision dataset but also ubiquitously exists in most real-world data. The experimental results show that our reweighting class strategy can alleviate this problem and increase the top-5 accuracy by $1.1\%$.

Besides, with reweighting bag-specific instance saliency strategy, the model can pay more attention on the representative instances, which can increase the top-5 accuracy by $0.3\%$. The above three strategies focus on alleviate the data bias. To be specific, reweighting class strategy can alleviate the bias arising from unbalanced class sizes. Reweighting cluster strategy can alleviate the bias arising from unbalanced instance difficulty. Reweighting bag-specific instance saliency strategy can alleviate the bias arising from unbalance instance representativeness.

On the other hand, the other two strategies focus on the instances with noisy labels. With reweighting instance strategy, the model can pay more attention on the more confidential instances, and the influence of less confidential instances can be suppressed. With reweighting label strategy, we can alleviate the issue when an instance belongs to two classes at the same time. Benefiting from these two strategies, our model can be further improved by $0.9\%$ on top-5 accuracy.


\subsection{Final Results in the WebVision Challenge}
\label{sec:details}

\begin{table}
\caption{Effects of various tricks in the WebVison challenge.}
\label{tab:trick:result}
\begin{center}
\begin{tabular}{ @{\hspace{2mm}} l @{\hspace{5mm}} c @{\hspace{2mm}} c @{\hspace{2mm}} }
\hline \hline
Tricks  &  Top-5 accuracy  &  Improvement \\
\hline
Remove noise out of top 15  &  $77.0\%$ & $0.5\%$ \\
Model ensemble  &  $78.3\%$  &  $1.3\%$ \\
Multi-crop testing  &  $79.3\%$  &  $1.0\%$ \\
Multi-scale testing  &  $79.8\%$  &  $0.5\%$ \\
\hline \hline
\end{tabular}
\end{center}

\end{table}

\begin{table}[t]
\caption{The officially reported results of WebVision image classification challenge 2018.}
\label{tab:webvision:result}
\begin{center}
\begin{tabular}{ @{\hspace{2mm}} c @{\hspace{5mm}} l @{\hspace{5mm}} c @{\hspace{2mm}} }
\hline \hline
Rank & Team name & Top-5 Accuracy \\
\hline
1 & Vibranium (Our team) & ${\bf 79.25}\%$ \\
2 & Overfit & $75.30\%$ \\
3 & ACRV\_ANU & $69.56\%$ \\
4 & EBD\_birds & $69.44\%$ \\
5 & INFIMIND & $68.74\%$ \\
6 & CMIC & $61.14\%$ \\
\hline \hline
\end{tabular}
\end{center}

\end{table}

Besides our reweighting strategies, in the WebVision challenge, we also use some useful tricks to further improve the performance. The tricks and the corresponding effectiveness are show in Table~\ref{tab:trick:result}. The first trick is that we found that the instances from Flickr usually are unreliable. So, we only use the top $15$ instances from Flickr for each class. This trick can improve the performance of our final model by $0.5\%$ on top-5 accuracy.

The others are three common tricks, \ie~model ensemble, multi-crop testing, multi-scale testing. For model ensemble, we train $29$ models and ensemble their results. This trick can improve the top-5 accuracy by $1.3\%$. For multi-crop and multi-scale tricks, we resize and crop the input images multiple times and fuse the prediction results. These tricks can improve the top-5 accuracy by $1.0\%$ and $0.5\%$ respectively.

Benefiting from our URNet and some experimental tricks, we achieve the first place on the WebVision Image Classification Challenge 2018. As shown in Table.~\ref{tab:webvision:result}, our method achieves $79.25\%$ top-5 accuracy on final test data, which outperforms the second method by about $4\%$. The results verify the effectiveness of our URNet.



\section{Conclusion}
\label{sec:conclusion}

In this paper, we demonstrate that every instance has the potential to contribute positively by alleviating the noise and bias via reweighting the influence of different classes as well as their labels, large instance clusters, small instance bags and their confidence. In this manner, the influence of the noise and bias in the web data can be gradually alleviated, which leads to the steadily improving performances of URNet. Experimental results in the WebVision 2018 challenge with 16 million noisy training images from 5000 classes show that our approach outperforms several state-of-the-art models and ranks the first place in the image classification task.

\ifCLASSOPTIONcompsoc
  \section*{Acknowledgments}
\else
  \section*{Acknowledgment}
\fi
This work was supported in part by National Natural Science Foundation of China (No.61672072), Beijing Nova Program (No. Z181100006218063) and Fundamental Research Funds for the Central Universities.

\ifCLASSOPTIONcaptionsoff
  \newpage
\fi



\bibliographystyle{IEEEtran}
\bibliography{URNetbib}

\begin{IEEEbiography}[{\includegraphics[width=1in,height=1.25in,clip,keepaspectratio]{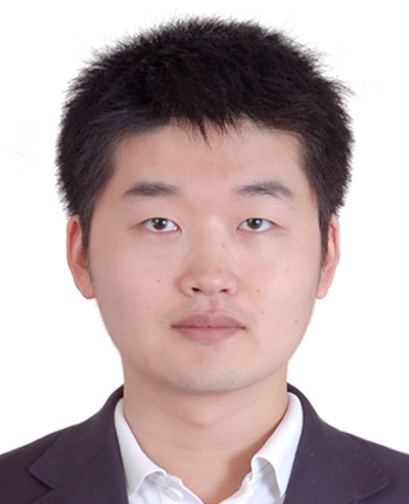}}]
{Jia Li} is currently an associate Professor with the School of Computer Science and Engineering, Beihang University, Beijing, China. He received the B.E. degree from Tsinghua University in 2005 and the Ph.D. degree from the Institute of Computing Technology, Chinese Academy of Sciences, in 2011. Before he joined Beihang University in Jun. 2014, he servered as a researcher in several multimedia groups of Nanyang Technological University, Peking University and Shanda Innovations. He is the author or coauthor of over 50 technical articles in refereed journals and conferences such as TPAMI, IJCV, TIP, CVPR, ICCV and ACM MM. His research interests include computer vision and multimedia big data, especially the learning-based visual content understanding. He is a senior member of IEEE and CCF.
\end{IEEEbiography}

\vspace{-1cm}

\begin{IEEEbiography}[{\includegraphics[width=1in,height=1.25in,clip,keepaspectratio]{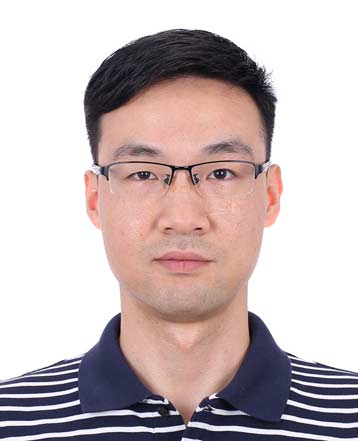}}]
{Yafei Song} is currently a postdoc researcher with the School of Electronics Engineering and Computer Science, Peking University, Beijing, China. He received the B.E. degree from Beijing Institute of Technology in 2010 and the Ph.D. degree from Beihang University in 2017. His research interests include computer vision, machine learning, augmented reality, and robotics.
\end{IEEEbiography}

\vspace{-1cm}

\begin{IEEEbiography}[{\includegraphics[width=1in,height=1.25in,clip,keepaspectratio]{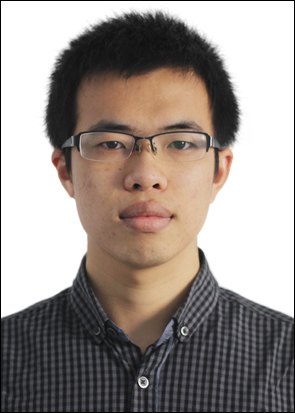}}]
{Jianfeng Zhu} is a research engineer in Computer Vision Technology Department of Baidu. His research interests include object detection and recognition, multi-modal learning and image retrieval.
E-Mail: zhujianfeng03@baidu.com
\end{IEEEbiography}

\vspace{-1cm}

\begin{IEEEbiography}[{\includegraphics[width=1in,height=1.25in,clip,keepaspectratio]{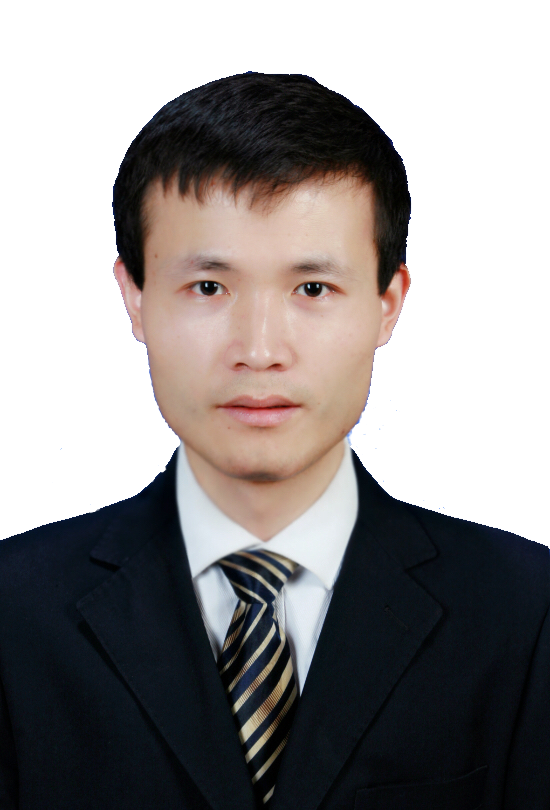}}]
{Lele Cheng} is a senior algorithm engineer in Computer Vision Technology Department of Baidu. His research interests include large-scale image recognition, image retrieval,face recognition.
E-Mail: chenglele@baidu.com
\end{IEEEbiography}

\vspace{-1cm}

\begin{IEEEbiography}[{\includegraphics[width=1in,height=1.25in,clip,keepaspectratio]{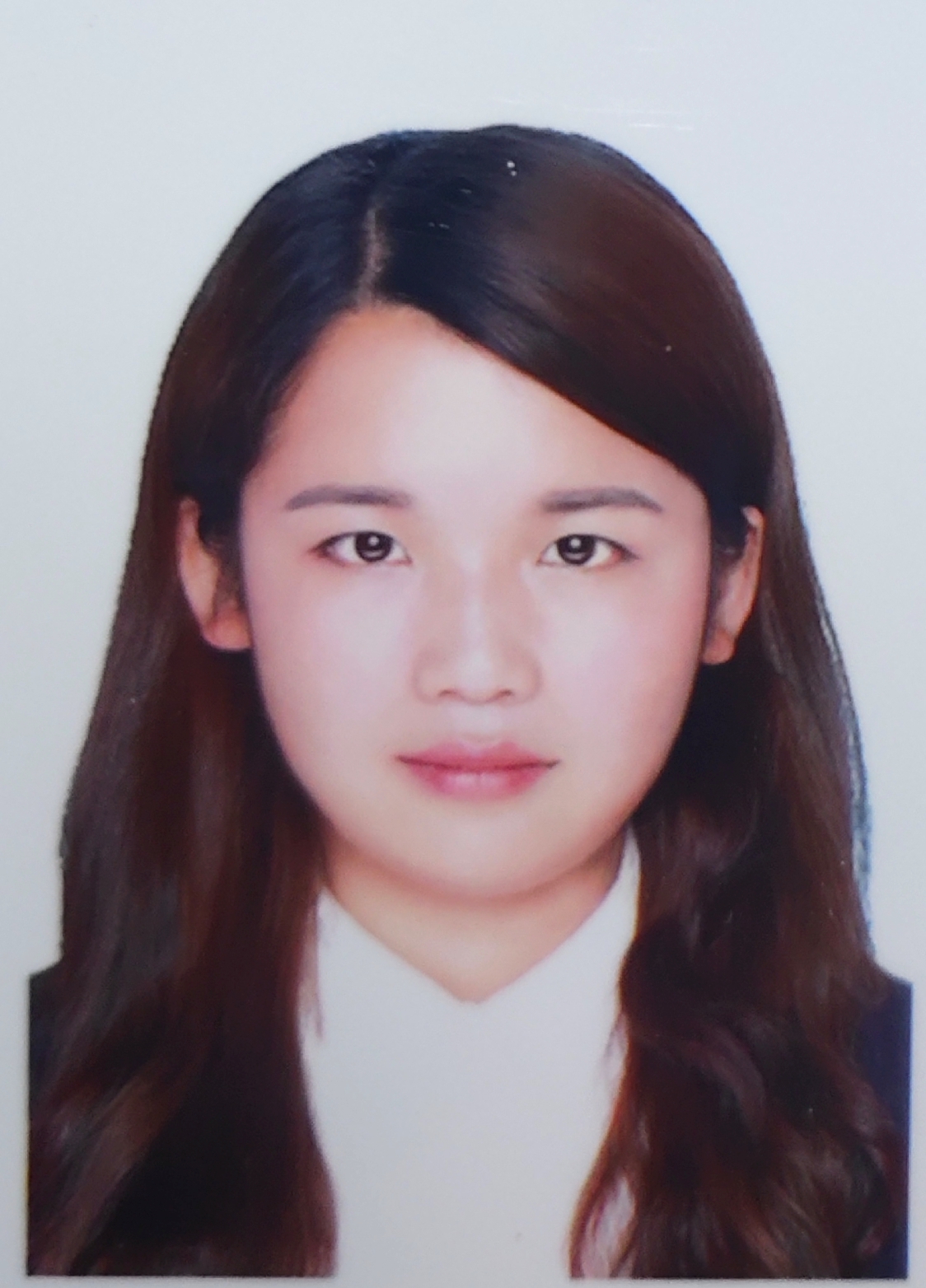}}]
{Ying Su} is a research engineer in Computer Vision Technology Department of Baidu. Her research interests include object detection and recognition, natural language processing.
E-Mail: suying02@baidu.com
\end{IEEEbiography}

\vspace{-1cm}

\begin{IEEEbiography}[{\includegraphics[width=1in,height=1.25in,clip,keepaspectratio]{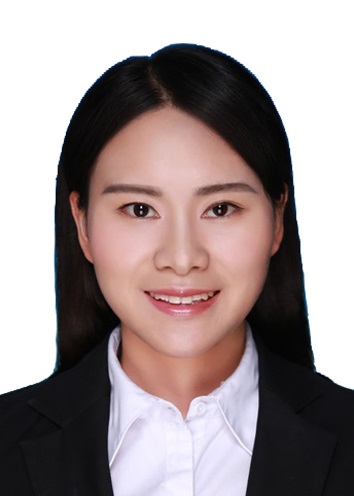}}]
{Lin Ye} is an engineer from Baidu responsible for image processing. She graduated from the Institute of Computing Technology of the Chinese Academy of Sciences in January 2018, majoring in the structure of computer systems. Currently, her focus is on mobile image classification and image caption.
E-Mail: yelin02@baidu.com.
\end{IEEEbiography}

\vspace{-1cm}

\begin{IEEEbiography}[{\includegraphics[width=1in,height=1.25in,clip,keepaspectratio]{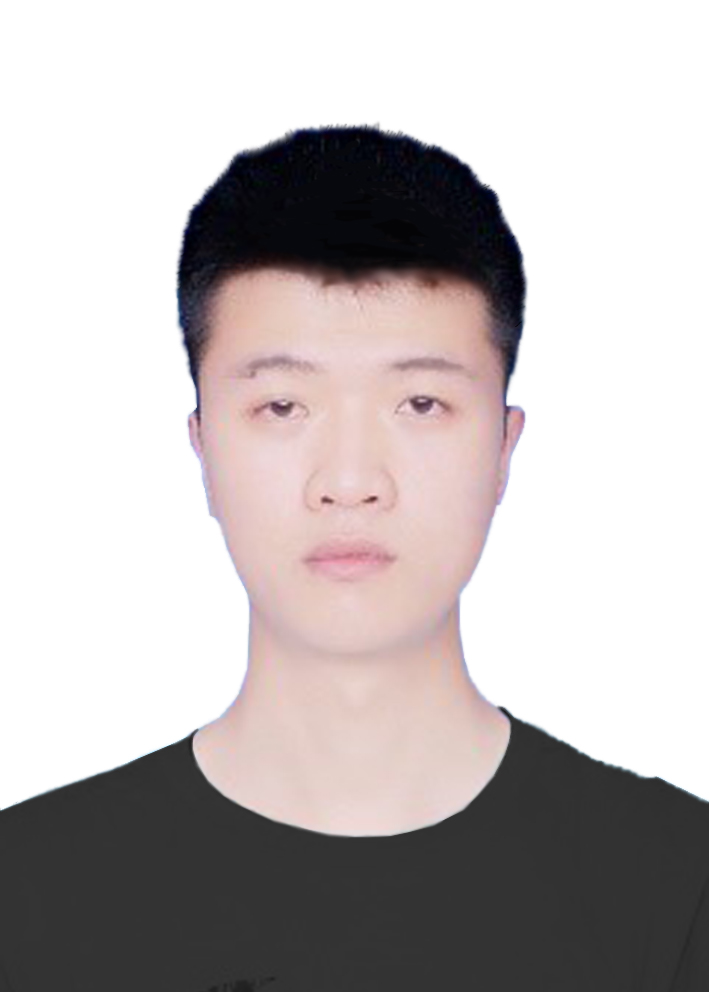}}]
{Pengcheng Yuan} is currently pursuing the Master degress with the State Key Laboratory of Virtual Reality Technology and Systems, School of Computer Science and Engineering, Beihang University. His research interests include computer vision and machine learning.
E-mail: yuanpengcheng@buaa.edu.cn.
\end{IEEEbiography}

\vspace{-1cm}

\begin{IEEEbiography}[{\includegraphics[width=1in,height=1.25in,clip,keepaspectratio]{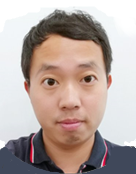}}]
{Shumin Han} is a research engineer in Computer Vision Technology Department of Baidu. His research interests include object detection and recognition, multi-modal learning and image retrieval.
E-Mail: hanshumin@baidu.com.
\end{IEEEbiography}

\end{document}